# Neural Computing for Online Arabic Handwriting Character Recognition using Hard Stroke Features Mining

**Amjad Rehman**

Artificial Intelligence and Data Analytics (AIDA) Lab. CCIS Prince Sultan University Riyadh KSA
School of Computing Universiti Teknologi Malaysia, Malaysia

**Abstract:** Online Arabic cursive character recognition is still a big challenge due to the existing complexities including Arabic cursive script styles, writing speed, writer mood and so forth. Due to these unavoidable constraints, the accuracy of online Arabic character's recognition is still low and retain space for improvement. In this research, an enhanced method of detecting the desired critical points from vertical and horizontal direction-length of handwriting stroke features of online Arabic script recognition is proposed. Each extracted stroke feature divides every isolated character into some meaningful pattern known as tokens. A minimum feature set is extracted from these tokens for classification of characters using a multilayer perceptron with a back-propagation learning algorithm and modified sigmoid function-based activation function. In this work, two milestones are achieved; firstly, attain a fixed number of tokens, secondly, minimize the number of the most repetitive tokens. For experiments, handwritten Arabic characters are selected from the OHASD benchmark dataset to test and evaluate the proposed method. The proposed method achieves an average accuracy of 98.6% comparable in state of art character recognition techniques.

## 1. Introduction

Offline handwriting recognition is quite different from online [1-5]. Nonetheless, there are some promising commercial products to recognize online handwriting, in particular for Latin and Chinese languages such as PenReader®, ritePen®, and Calligrapher®. However, Arabic script recognition, accuracy is still lacking for both online and offline scripts due to complexities of writing, in the isolated and cursive script both. The research interests in online Arabic script recognition have grown significantly and it is an open field in pattern recognition and image processing applications [6-10].

The most recent recognition systems are recognizing the handwritten text in a special style [11]. Two core types of recognition systems are used- i) specific language-based system in which training is performed on one selected language and then select one of them for prediction [12] ii) a unified system which at least trained on multiple languages. Due to the nature of challenges in distinguishing among different handwriting styles, uncertainty types of human writings, different shapes, and sizes of the characters, and individual writing conditions, it is particularly difficult to find a stable invariant feature set. Some studies have aimed specifically at segmenting handwriting patterns into some strokes and extracting all features from these strokes directly [13]. To investigate the properties of an alphabet, the features could be divided into two categories name global and local features. All of these features are separated into three groups such as structural features, statistical features and their fusion [13]. The structural and statistical features set are extracted based on the geometrical and topological aspects of the stroke sequence [14,15]. Various computer-based techniques are introduced which followed the preprocessing to strong extraction in the form of features, and finally classification through Support Vector Machines, Neural Networks, etc [16-24].

In this work, an improved online system is designed for Arabic cursive character recognition using hard stroke features. The proposed system involves key steps: preprocessing of original data, character segmentation; feature extraction and classification. The major contributions are:

i) A strokes based segmentation is performed in which each character can illustrate the own direction-length. Because the framework of a stroke helps to estimate the direction-length. Therefore, based on the minimum and/or maximum values of horizontal (x) and vertical (y) coordinate points, it is decided whether the direction-length is vertical or horizontal.

ii) In the feature extraction step, the set of structural and statistical features is computed based on the shape, distance analysis, and the ratio of connected components to other parts of segmented strokes respectively. These calculated features are combined and fed to a multilayer neural network for final recognition.

The remaining paper is organized as such: Section 2 described the Related Work in which existing techniques are presented. The proposed approach is presented under section 3 which follows the section 4 as data collection and results. The last section presents the conclusion and future work.

## 2. Background

A lot of techniques are reported for automatic segmentation and recognition of Latin script and individual characters for both online and offline. However, a few are reported on Arabic script recognition [4, 25,26]. Abed and Alasad [27] extracted features of Arabic characters using zoning techniques and claimed 93.61% accuracy by using Error Back Propagation Artificial Neural Network (EBPANN) as a classifier. Ismail and Abdullah (2012), proposed a rule-based classifier for online Arabic character recognition and used horizontal and vertical projection profiles features. The reported accuracy was 97.6%.

Harouni et al., [28] proposed an online Arabic character recognition strategy using BP/MLP with geometric features set. The reported accuracy was 100% without standard dataset employed.

Likewise, Harouni et al [29] performed Persian/Arabic online character recognition using artificial neural networks and particle swarm optimization (ANN–PSO) classifier. 88.47% accuracy reported on TMU dataset. Ramzi and Zahary [30], also recognized online Arabic characters on a database of online and offline character samples (1,050 samples for training and 420 for testing) collected from five users. Chain code online features combined with geometric features were extracted and 74.8% recognition accuracy attained using a backpropagation neural network.

Al-Helali and Mahmoud [31] presented a statistical framework for online Arabic character recognition. They improved the recognition accuracy by using delayed strokes at the different phases differently and statistical features of segmented characters of the Online-KHATT database.

Saba [32] classified Arabic script by using a fuzzy ARTMAP classifier with a set of statistical features. Experiments were conducted on the IFN/ENIT database and an accuracy of 94.72% reported. Based on the complexities of Arabic cursive handwriting styles, speed, touching, overlapping and inherent properties, a pre-processing strategy is needed to collaborate with intelligent techniques to enhance the accuracy such as neural networks [33-37], GA [38-40], SVM [41-43]. However, character boundaries detection in touched and overlapped consecutive characters made this issue more crucial. Consequently, a few segmentation free techniques for script recognition are also reported termed as holistic approaches but worked on a small dataset only and training/testing of intelligent techniques is another issue [44]. Lastly, the cursive segmentation methods affect directly the efficient reliability analysis of recognition and the performances of the different existing handwriting techniques are significantly low for Arabic script recognition [4, 45]. Therefore, a systematic and efficient segmentation scheme is desired to separate a cursive Arabic word into its characters precisely.

## 3. Proposed Method

An improved online system is proposed for online Arabic cursive character recognition using hard strokes features set. The proposed system consists of three core steps including preprocessing of original data, character segmentation, strokes feature extraction and classification. The detailed description of each step detailed below.

### 3.1. Preprocessing

In Arabic handwriting main stroke, body characteristics exist in their common alphabet. These are some minor differences among them that may differentiate using small diacritical marks above and below the alphabets. Preprocessing is normally desired to eliminate unnecessary detail [46-48]. Hence a hybrid interpolation and smoothing method for the raw data is proposed to reconstruct a compact-looking of the handwriting patterns without missing and transforming the inherent character-writing structure. Initially, let *P* signifies the real-time trajectory points of the raw data pattern, i.e. the set of points {(i, j)}, *i* and *j* are the *x-direction* and *y-direction* respectively and the maximum *value of i* and *j* is N; each alphabet can be partitioned into disjoint non-empty subsets, i.e. strokes $S_1, S_2, \ldots, S_n$, which could be presented and recorded as shown in Equation (1).

$$P = \Upsilon_{i=1}^{n} S_i \tag{1}$$

where n is the total number of strokes within a handwriting pattern. To apply the proposed interpolation process, all X, Y coordinate points are obtained separately using Equation (2) and (3).

$$X_{RHP} = \Upsilon_{i=1}^{N} x_i \tag{2}$$

$$Y_{RHP} = \Upsilon_{i=1}^{N} y_i \tag{3}$$

where *N* is the total number of coordinate points of input handwritten pattern.
The following equations are applied to smooth and interpolate raw input pattern, which is amended and altered from [49].

$$x_i = \frac{3}{5} p'(x_{i-1}) + \frac{1}{5} p(x_i) + \frac{1}{5} p(x_{i+1}) \tag{4}$$

$$y_i = \frac{3}{5} p'(y_{i-1}) + \frac{1}{5} p(y_i) + \frac{1}{5} p(y_{i+1}) \tag{5}$$

where the raw input coordinate point of x and y values will be replaced with the interpolated values at the point of $P_i$; and the $p'_i$ is the interpolated coordinate point of *ith* interpolated point.

### 3.2. Segmentation

Segmentation is an important step in many image processing applications such as optical character detection, medical imaging, and agriculture, etc [51-55]. In a handwritten segmentation-based strategy, the main challenge is how to reduce the number of over-segmented handwriting patterns and also to guarantee the correct segmentation of the character boundaries. The first stroke in each character could illustrate the own direction-length of the handwritten character, and also the framework of a stroke helps to estimate the direction-length. Therefore, based on the minimum and/or maximum values of x and y coordinate points, the Equations (6), (7) and (8) decide whether the direction-length is vertical or horizontal.

$$S_{length} = \bigcup_{i=1}^{n} S_i^{length} \quad (6)$$

$$S_i^{length} = (x_{max_i} - x_{min_i}) - (y_{max_i} - y_{min_i}) \quad (7)$$

$$S_i^{direction-length} = \begin{cases} Horisontal\ Length\ Format & S_i^{length} \geq 0 \\ Vertical\ Length\ Format & Otherwise \end{cases} \quad (8)$$

$S_{length}$ is the union of all length strokes, i.e. $S_1, S_2, \ldots, S_n$, where *n* is the total number of strokes in a given handwriting pattern. Moreover, $S^{direction-length}$ denotes the direction length of each stroke as well. All critical points, i.e. all maximum and/or minimum (x, y) coordinate points, of each stroke are extracted based on direction length; for example, if the direction-length is horizontal length format, then the critical points are distributed on the x-axis. The proposed algorithm shows how to detect and save all critical points of the strokes in horizontal length format into a stroke.

The main idea of segmentation is based on the distribution of points on both sides of a local-maximum value pixel; this helps to provide an inherent information structure of handwriting patterns and reduce the effect of the sudden change in direction and/or extremely short breaks. Hence, the proposed method could consider the better shape of individual handwriting strokes and their spatial relations. A pixel could be considered a local-maximum value pixel if there exists an ascending sequence of y-values (Y-axis) on the right side and a descending sequence of y-values (Y-axis) on the left side of the pixel with a distribution ratio of 0.05×N. more detail is shown in Equation (9).

$$f_i(y) \leq f_{i+1}(y) \leq f_{i+2}(y) \leq \ldots \leq f_m(y) \quad (9)$$

where m is 0.05 percent of x-axis length, i.e. 0.05×N; the proposed algorithm absolutely supports different writing styles of Arabic script, and also can separate each stroke into some parts termed as tokens to extract appropriate features.

### 3.3. Features extraction and classification

Features extraction and selection is a significant stage in the whole process. Too many features confuse and overburden the classifier while too few features reduce the recognition accuracy [56,57]. Sometimes only few discriminative features are enough to recognize patterns/characters [58,59,60]. Therefore, a set of structural and statistical features is extracted that is based on the shape, distance analysis and the ratio of connected components to other parts of segmented strokes respectively. Moreover, it has been mentioned in previous sections, each character has some strokes, and also by applying the proposed process, each stroke will be divided into some tokens, as shown in Equations (1) and (10). Based on these tokens, these features are extracted, which are enough for recognizing online Arabic handwritten alphabets.

$$S = \bigcup_{i=1}^{m} T_i \quad (10)$$

where *m* is the total number of tokens inside each stroke, T is a symbol of Token and denotes some disjoint non-empty subsets of coordinate points of its stroke.

The first feature calculates the length ratio of each token to its stroke, see Equations (11) and (12), where four categories are defined as shown in Table 1.

$$T_i^{length} = (x_{max_i} - x_{min_i}) - (y_{max_i} - y_{min_i}) \tag{11}$$

$$f_{lengthRatio}^{token_i} = \frac{T_i^{length}}{S^{length}} \times 100 \tag{12}$$

**Proposed Algorithm**

**Step 1-** Input: $X_{RHP} = \bigcup_{i=1}^{N} x_i$ and/or $Y_{RHP} = \bigcup_{i=1}^{N} y_i$

**Step 2-** Focus on the first stroke.

**Step 3-** Extract the maximum and minimum of X, Y coordinate points of the stroke to compute its framework.

**Step 4-** Detect the direction-length of Input Handwritten Pattern (IHP) by using the stroke's framework, i.e. horizontal length or vertical length format.

**Step 5-** Detect local maximum point: hereupon if the IHP length is horizontal format then go to 6, otherwise go to 7.

**Step 6-** Assume that $f_m(x, y)$ is a Local Maximum Point of the stroke, if and only if Y values are dependent on the distribution of points on both sides of $f_m(x, y)$; which are greater than the minimum length of 0.05×N; hence there exists the local maximum point on X-axis, i.e. horizontal axis, at $f_m(x, y)$ point; go to 8.

**Step 7-** Assume that $f_m(x, y)$ is a Local Maximum Point of the stroke, if and only if X values are dependent on the distribution of points on both sides of $f_m(x, y)$; that are greater than the minimum length of 0.05×N; hence there exists the local maximum point on Y-axis, i.e. vertical axis, at $f_m(x, y)$ point; go to 8.

**Step 8-** Output: Save each $f_m(x, y)$ as a detected critical point of the stroke

**Step 9-** Find the next stroke; if so, go to 3; otherwise, return all detected critical points that belong to the stroke.

Table 1: Categorizing length ratio of the tokens in terms of their strokes

| Length    | Short  | Middle-short | Middle-long | long     |
|-----------|--------|--------------|-------------|----------|
| Ratio (%) | (0,25) | (25,50)      | (50,75)     | (75,100) |

The next feature is to estimate each token direction inside its stroke represented in Equation (13).

Both the following Equations (14) and (15) demonstrate how the orientation of each token is symbolized.

$$f_{direction}^{token_i} = \tan^{-1}\left(\frac{y_{max_i} - y_{min_i}}{x_{max_i} - x_{min_i}}\right) \tag{13}$$

$$f^{token_i}_{midPoint} = P\left(\frac{x_{max_i} + x_{min_i}}{2}, \frac{y_{max_i} + y_{min_i}}{2}\right) \tag{14}$$

$$S^{token_i}_{orientation} = \begin{cases} OnClockwise & f^{token_i}_{midPoint} \geq The\,midpoint\,of\,the\,token \\ OnCounterClockwise & Otherwise \end{cases} \tag{15}$$

For the classification, all characters with the same number of strokes are clustered into a given group, as presented in Table 2. Therefore, these groups can help to gain better clustering of the features of the handwriting patterns as the desired input set to the classifier.

The next step is to define a highly appropriate classifier for final recognition. For this purpose, we utilized a backpropagation multilayer perceptron (BP/MLP) neural network detailed in the next section.

### 3.4. Neural computing for characters recognition

The neural network has confirmed to be a better competitor upon up-to-date techniques to obtain the best accuracy for complex real-time problems. Neural Network mimics natural human brain processing represented as neurons. These neurons take input in the form of features and transfer to other neurons for the learning process. In this work, a backpropagation learning algorithm with a multilayer perceptron is employed for the recognition process. The MLP with backpropagation includes a sandwich of the hidden layer (HL) among input and output layers. The selection of hidden layers and the number of neurons in the hidden layers is a critical task and problem-dependent; therefore in this work, we defined the selection process as follows:

$$Y_l = \delta_1\left(\sum_{j=0}^{P} w^Y_{lj}\left(\delta_h\left(\sum_{i=1}^{Q} w^h_{ji} x_i\right)\right)\right) \tag{16}$$

where, $w^Y_{lj}$ denotes the weights from the neuron j in the HL denotes by $h$, $w^h_{ji}$ denote the weight of $i$ neuron, $x_i$ is input layer data, and $\delta_1$ and $\delta_h$ denote the activation function of output and HL, respectively. Finally, the cost function is defined which minimized the data as follows:

$$\Phi = \frac{1}{2}\sum_{l=1}^{P}(Y_l - E_l) \tag{17}$$

$$\Phi = \frac{1}{2}\sum_{l=1}^{P} e_l^2 \tag{18}$$

where $\Phi$ denotes the minimized cost function, $Y_l$ is actual output and $E_l$ is expected output, respectively.

## 4. Experimental Results and Analysis

### 4.1. Dataset

In comparison to the offline handwriting, online handwriting has an almost limited standard dataset. However, for a fair comparison, OHASD Online Handwritten Arabic Sentence Database is employed [61]. OHASD composed of samples of paragraphs of complete sentences of 15 to 46 words each. Moreover, 154 paragraphs are containing 3,825 words and 19,467 characters written by 48 writers after excluding illegible and erratic handwriting. Table 2 presents the clustering of each Arabic alphabet based on shape and number of strokes.

Table 2: Grouping and clustering each alphabet in terms of its body shape and the number of strokes respectively.

| G.* 1 | G. 2 | G. 3 | G. 4 | G. 5 | G. 6 | G. 7 | G. 8 | C.** 1 | C. 2 | C. 3 | C. 4 |
|---|---|---|---|---|---|---|---|---|---|---|---|
| | | | | | | | | *1-stroke* | *2-stroke* | *3-stroke* | *4-stroke* |
| ا آ | ب ت ث | ج ح خ | ذ ر ز ژ | س ش ص ض | ط ظ ع غ | ف ق | ا ل م ن و س ه ی ع | ا ح د ر و ص | آ ب ج خ ذ | ز ض ط ظ ق غ ف گ ک | ب ث چ ژ ش |

*Group
**Cluster

### 4.2. Recognition Results

It is always hard to compare results in state of art techniques due to different datasets used. However, the performance of the proposed approach is computed using accuracy measures detailed below:

$$Accuracy = \frac{TP+TN}{TP+TN+FP+FN} \quad (19)$$

$$Recall = \frac{TP}{TP+FN} \quad (20)$$

$$Precision = \frac{TP}{TP+FP} \quad (21)$$

$$FNR = 1 - Recall \quad (22)$$

where TP denotes the correctly predicted images and FN denotes the incorrectly predicted samples. All the simulations are performed on MATLAB 2017 using a Personal Desktop system of 8GB of RAM and 4 GB graphics card.

To compute the proposed recognition results, the dataset is divided into training and test samples. For this purpose, 70% of collected images are utilized for training the classifier and the remaining 30% is utilized for testing the proposed approach. All testing results are computed by k fold cross validation where k=10. The recognition results are given in Table 3. From Table 3, 98.6% accuracy is achieved by employing k=10 fold validation whereas the other measures such as recall rate are 97.94%, the precision rate is 98.2%, and FNR is 2%, respectively. Similarly, for K=5, the achieved accuracy is 97.4% with FNR is 2.6%, respectively. From results, it is clear for K=10 achieves better performance.

Table 3: Proposed recognition performance using different K-folds

| Method | K-fold | | Performance measures | | | |
|---|---|---|---|---|---|---|
| | 5 | 10 | Recall rate (%) | Precision rate (%) | Recognition accuracy (%) | FNR (%) |
| **Neural Network** | ✓ | | 97.20 | 97.5 | 97.4 | 2.6 |
| | | ✓ | **97.94** | **98.2** | **98.6** | **2.0** |

Besides, a Monte Carlo simulation to authenticate the results in terms of consistency is computed. The results are shown in Table 4 for different iterations. From Table 4, it is described that the maximum achievable accuracy is 98.2% when several iterations (NoI) =100 and minimum achieve accuracy is 97.04% when NoI=500. The results show that the proposed method accuracy is a little bit changed when it is iterated for maximum time.

Table 4: Monte Carlo based recognition accuracy in the form of minimum, average, and maximum

| Method | Number of iterations | | | | | Recognition performance |
|---|---|---|---|---|---|---|
| | 100 | 200 | 300 | 400 | 500 | |
| BP/MLP | ✓ | | | | | **97.63,97.80,98.2** |
| | | ✓ | | | | 97.60,97.74,97.98 |
| | | | ✓ | | | 97.42,97.66,97.90 |
| | | | | ✓ | | 97.28,97.54,97.84 |
| | | | | | ✓ | 97.04,97.42,97.80 |

### 4.3. Discussion and Analysis

The proposed algorithm exhibits promising results to detect critical points from a set of handwriting patterns. It helps segment well-established patterns. Based on the overall percentages of the number of tokens happened within each character, the net result discloses several points. First, a prime target of improving this algorithm is to achieve the minimum intensive and a fixed number of tokens. Second, the minimum number and the most repetitive of tokens. In most characters, it's extremely happening as their number of strokes with considering dots; therefore, it is confident that the proposed algorithm will be able to obtain a reasonably accurate account on extracting desired patterns used in feature extraction and classification steps.

Generally, the purpose of the pre-processing step is to prepare the desired dataset from an input raw data of the handwriting image and is two-fold. First a standard task for exploiting the characteristic of the data to be analyzed; second, to get the desired dataset, without misshaping and transforming the intrinsic structure of the handwritten alphabets from the raw data as shown in Figure 1. Besides, handwriting segmentation step plays a vital role in applications related to the recognition system. The reasonable segmentation errors can be considered as a reliable reference for the textual feature extraction. Due to the high variability in the handwriting style, which causes some tokens of the cursive character strings may be overlapped and increases the difficulties in the segmentation of the correct boundaries of them, the proposed segmentation process is applied for preparing a set of suspicious segmentation pixels as potential token boundaries in a real sense as shown in Figure 2 and Figure 3

| Raw Input Image | Pre-processed Image | Raw Input Image | Pre-processed Image | Raw Input Image | Pre-processed Image |
|---|---|---|---|---|---|

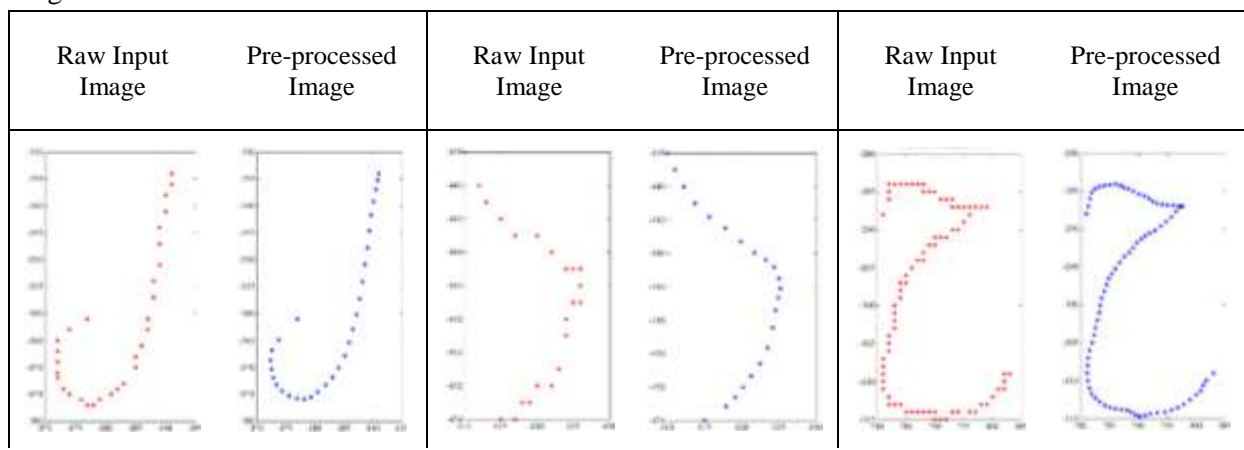

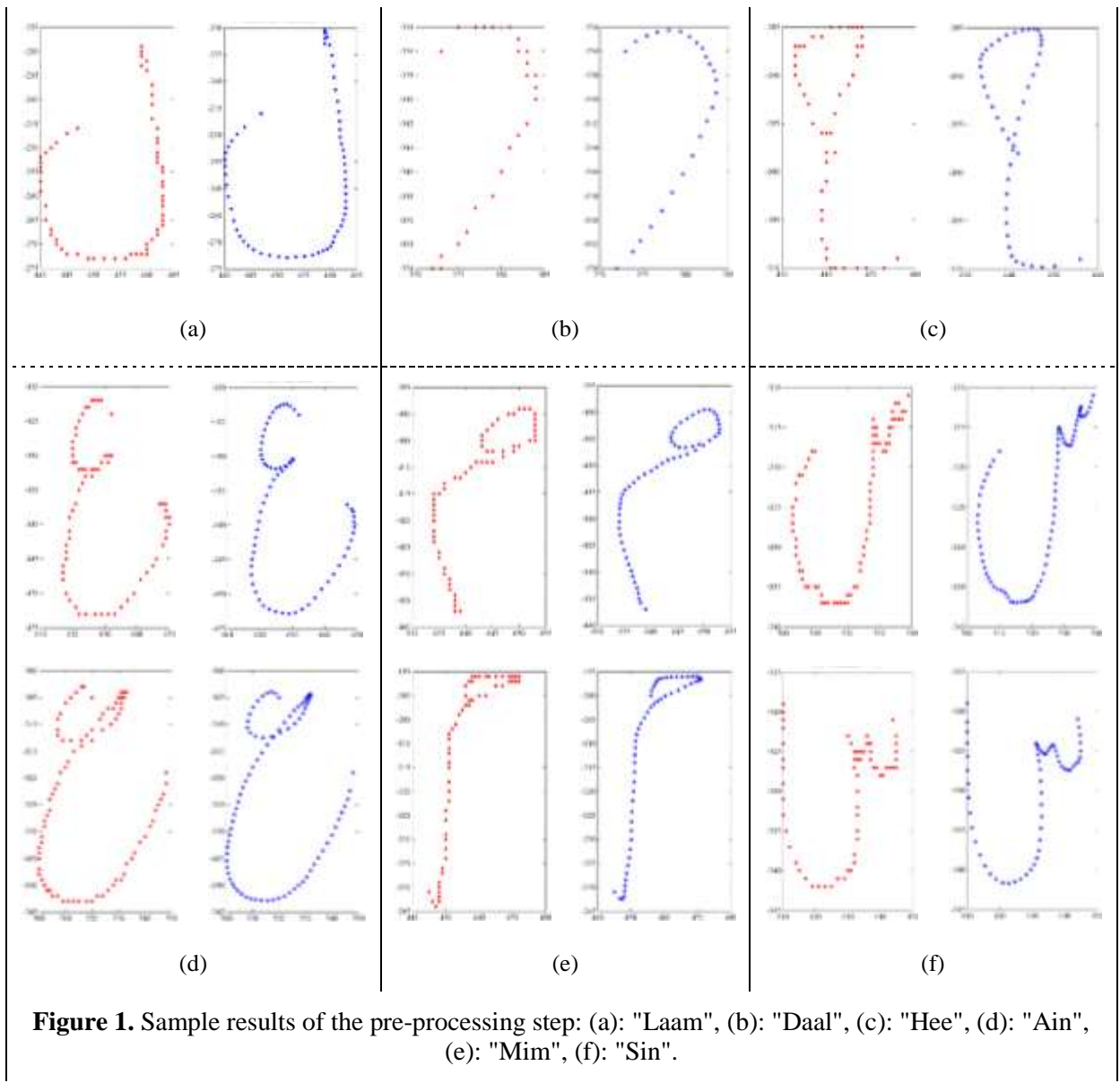

**Figure 1.** Sample results of the pre-processing step: (a): "Laam", (b): "Daal", (c): "Hee", (d): "Ain", (e): "Mim", (f): "Sin".

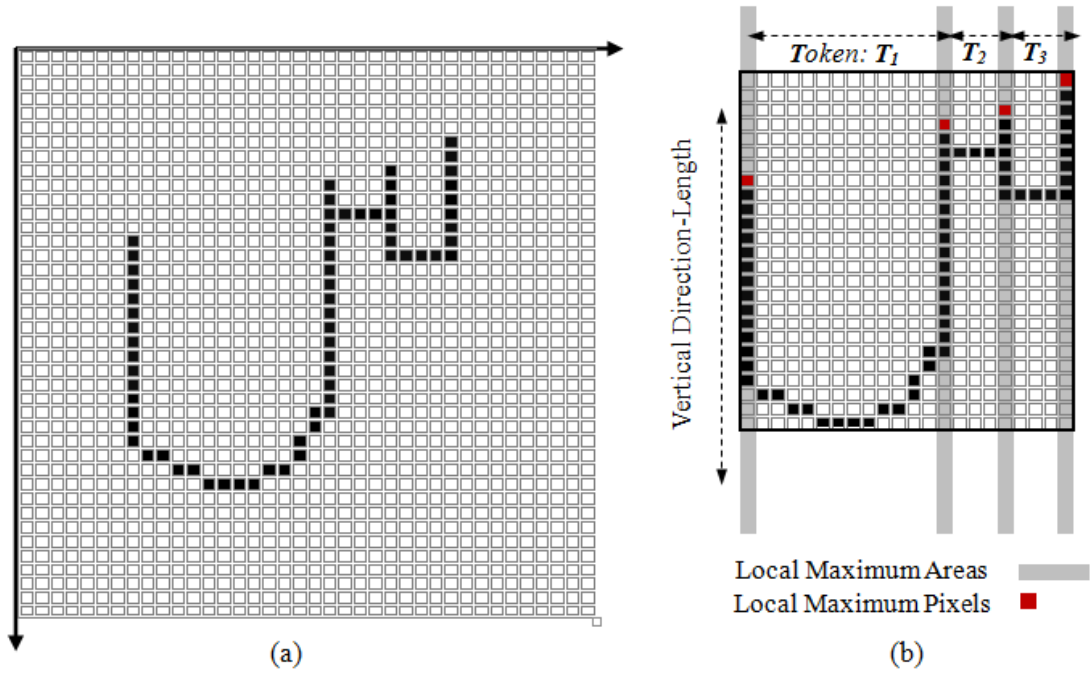

**Figure 2.** Handwriting pattern segmentation process: (a): a pre-processed handwriting pattern (b): character segmentation into individual parts (tokens) based on local-maximum value pixel.

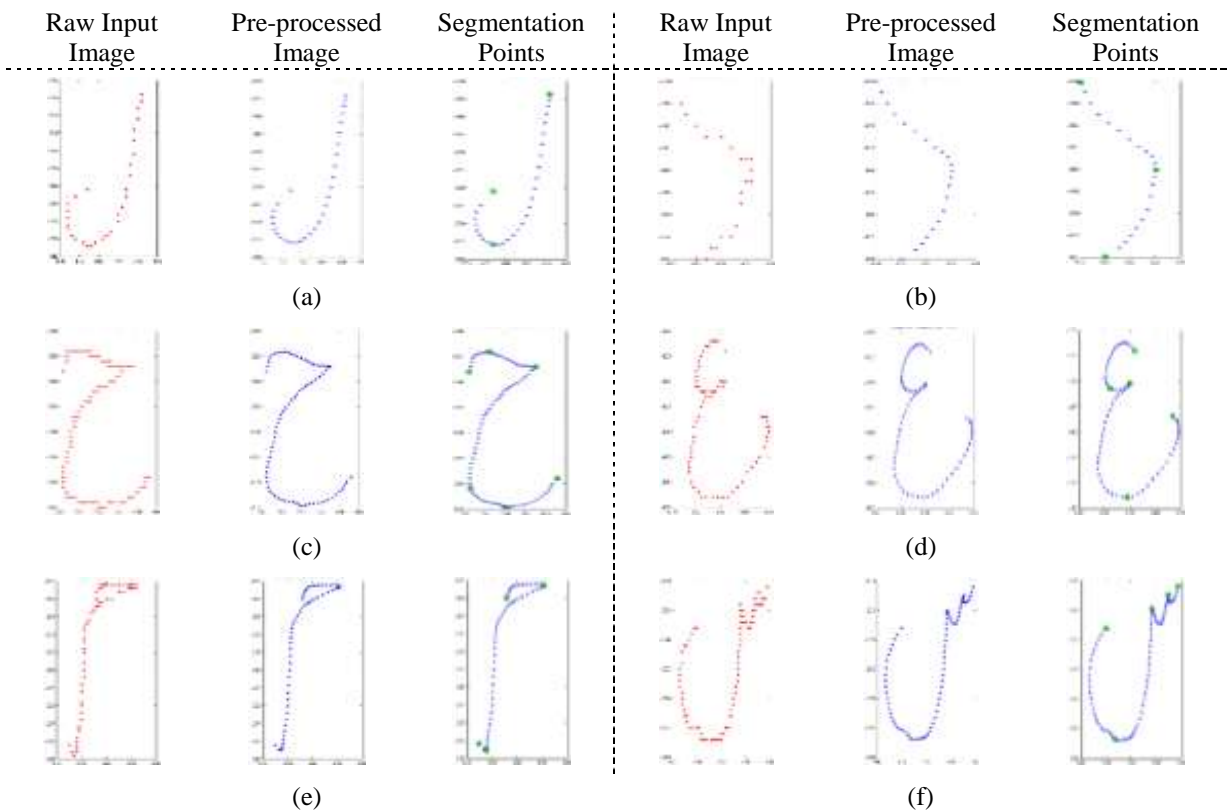

**Figure 3.** Sample segmentation results: (a): "Laam", (b): "Daal", (c): "Hee", (d): "Ain", (e): "Mim", (f): "Sin".

A sample feature extraction result could be seen in Figure 4. All extracted features are swapped to binary value as the same formats and styles for the data inputs of neural networks. The output of the classifier would be the defined classes that are belonged to all possible segmented characters.

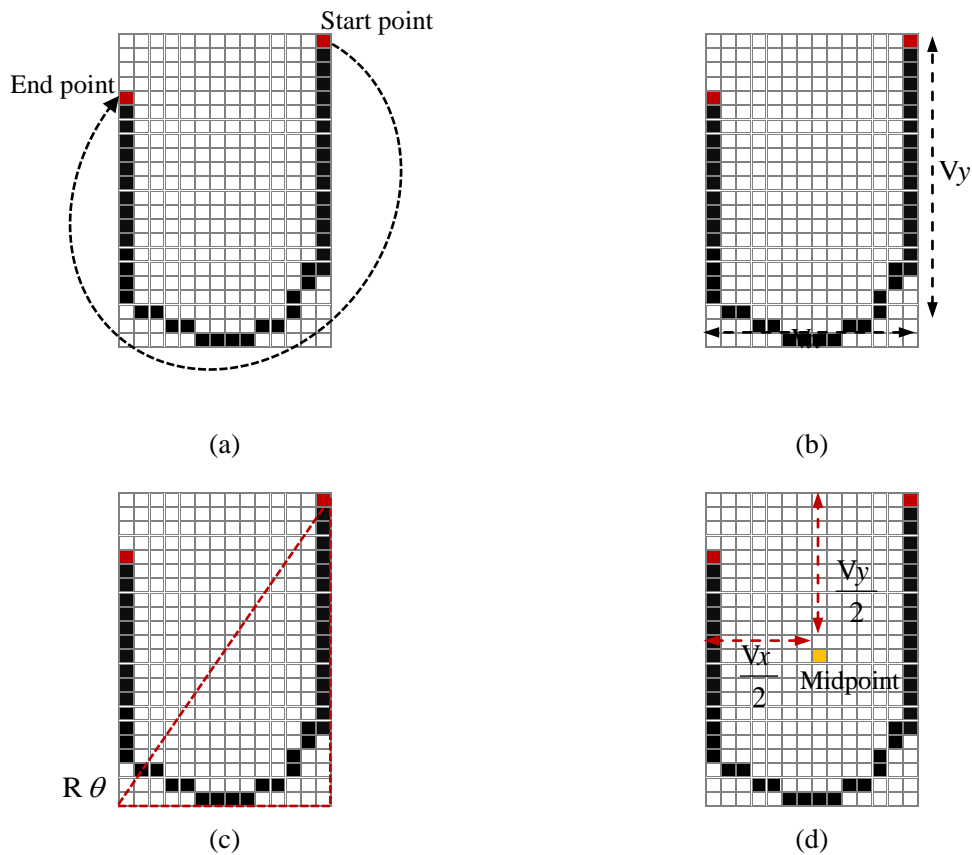

**Figure 4:** Feature extraction for each token: (a): token's orientation, (b): token's length, (c): token's direction, (d): token's midpoint.

Through the experiments, a discussion is presented about the inaccurate classification. It is shown that a few characters are misclassified due to similar appearance to their writing styles in different position shapes. Hence these characters cause considerable confusion for the character neural network. Besides, there exist a few characters, such as "Dal" and "Reh", with a striking resemblance to their main body characters, i.e. in the time of writing, the proposed algorithm does not face over-segmentation and can segment and distinguish them accurately. It is worst noted that the initial position shapes of character "Sad" are a close resemblance to the initial position shape of the character "Seen", which is written with incorrect jagged its vertical lines, despite using the feature of segmented character strokes, it is found that these characters are confused somewhat to each other. The main body and the second delayed strokes of some handwriting alphabet patterns together can provide more detailed information to extract a useful feature set for recognizing the characters, e.g. "Seen" vs. "Sheen", "Ain" vs. "Ghain". The most obvious finding to extract mentioned features from these tokens is that the number of features depends directly on the number of tokens within each character. Hence the maximum tokens signify in the status of features and help to get a fixed number input using in the classification design. So, all the mentioned features are extracted from these tokens and the classifier is trained and tested.

The outstanding merit of the proposed algorithm is to concentrate on several tokens in the same character within the limited and normal range. In other words, Table 5 shows, total tokens obtained in each character are reduced into a meaningful number. It is observed that the minimum and most repetitive number of the tokens obtained is almost equal, except four alphabets for the proposed readout algorithm, i.e. "Alef": "ﺍ", "Jeem": "ج", "Ain": "ع", and "Yeh": "ى".

Table 5: The overall percentage of the number of tokens happened within each character

| Approach | Overall Percentage | | |
|---|---|---|---|
| | Minimum | Most repetitive tokens | Maximum |
| Proposed algorithm | 69% | 74% | 11% |

As mentioned earlier, BP/MLP is employed for classification with modified sigmoid function as an activation function. The initialization of weights is approximately presumed with equal magnitude and also limited in a minimal bound [62-65]. In Equation (23), these initial values are randomly generated in the range of (-0.2,0.2).

$$\sqrt{\eta/N} = |w_i(0)| \quad for\ i = 0,1,2,...,N \qquad (23)$$

Where $\eta$ is the learning rates, learning continued until an appropriate output is matched with the actual output nodes on the same input dataset. To preserve the generation of initial weights out of the given range $\eta$ must be selected very small [67-70], so here it is 0.1. As a final point, internal threshold and momentum coefficient values are selected 0.0 and 0.05 respectively; The BP/MLP classifier is utilized 70% of samples for training the network and remaining 30% for performing tests.

5. Conclusion

In this article, a robust method is proposed for detecting the desired critical points from the vertical and horizontal direction-length of online Arabic handwriting stroke features. the extracted stroke features are recognized through the BP/MLP neural network and achieved an accuracy of 98.6% using OHASD benchmark database. From the results, it is concluded that the preprocessing and segmentation steps assisted in finding the best strokes feature for the recognition of online Arabic characters. The evaluation of the system reveals that the proposed algorithm could support to find all meaningful critical points of the different styles and speeds of writing. Moreover, from the results, it is also concluded that the maximum number of iterations like Monte Carlo does not affect the accuracy of the system.
In the future, instead of strokes feature, a deep learning framework will be utilized to extract automatic features at large scale for precise classification of Arabic characters. Moreover, other languages like Persian, Urdu, Hindi cursive characters will also be explored.

**Acknowledgement**
This work was supported by Artificial Intelligence and Data Analytics (AIDA) Lab CCIS Prince Sultan University Riyadh Saudi Arabia. Author is thankful for the support.